\documentclass[10pt,twocolumn,letterpaper]{article}

\usepackage{cvpr}              
\usepackage{booktabs}
\usepackage{graphicx}

\definecolor{cvprblue}{rgb}{0.21,0.49,0.74}
\usepackage[pagebackref,breaklinks,colorlinks,allcolors=cvprblue]{hyperref}

\def\paperID{17965}
\def\confName{CVPR}
\def\confYear{2026}

\title{Exploring Fusion Strategies for Multimodal Vision-Language Systems}

\author{Regan Willis \\
University of South Carolina \\
Columbia, SC 29208 USA \\
{\tt\small rw11@email.sc.edu}
\and
Jason Bakos \\
University of South Carolina \\
Columbia, SC 29208 USA \\
{\tt\small jbakos@cse.sc.edu}
}

\begin{document}
\maketitle
\begin{abstract}
Modern machine learning models often combine multiple input streams of data to more accurately capture the information that informs their decisions. In multimodal machine learning, choosing the strategy for fusing data together requires careful consideration of the application’s accuracy and latency requirements, as fusing the data at earlier or later stages in the model architecture can lead to performance changes in accuracy and latency. To demonstrate this trade-off, we investigate different fusion strategies using a hybrid BERT and vision network framework that integrates image and text data. We explore two different vision networks: MobileNetV2 and ViT. We propose three models for each vision network, which fuse data at late, intermediate, and early stages in the architecture. We evaluate the proposed models on the CMU-MOSI dataset and benchmark their latency on an NVIDIA Jetson Orin AGX. Our experimental results demonstrate that while late fusion yields the highest accuracy, early fusion offers the lowest inference latency. We describe the three proposed model architectures and discuss the accuracy and latency trade-offs, concluding that data fusion earlier in the model architecture results in faster inference times at the cost of accuracy.
\end{abstract}    
\section{Introduction}
With the recent success and wide-adoption of artificial intelligence, researchers, businesses, and consumers are demanding higher and higher levels of intelligence from their systems. While specialized models are suitable for solving specific problems in constrained environments, there is a growing desire for our systems to reason about problems more intelligently. Multimodal systems can make our systems more robust to failure, more integrated with our experience, and capture valuable information that otherwise goes unseen.

Multimodal systems can be more robust to failure, as in autonomous driving \cite{khuang}. While an autonomous driving system that uses just Electro-Optical (EO) cameras will struggle in low light, a system that also includes IR cameras can continue to perform. Most autonomous driving systems are now multimodal in some form \cite{khuang}. There is also an increasing consumer demand for language models to be more easily integrated with the experience of users that has lead to a rise in vision-language models (VLMs) like CLIP, which allows users to leverage visual data as well as textual \cite{ghosh}. As we ask our models to solve more complex problems, it becomes essential that they have the capability to accept multiple data modalities.

\begin{figure}[t]
    \centering
    \includegraphics[width=\linewidth]{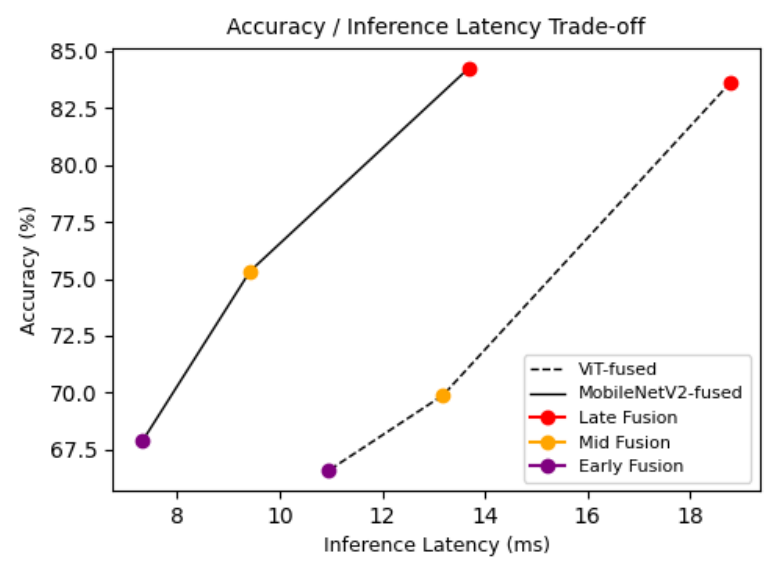}
    \caption{Accuracy vs. Inference Latency trade-off space for the proposed MobileNetV2-fused and ViT-fused models on the Jetson Orin AGX.}
    \label{fig:tradeoff}
\end{figure}

Multiple independent models can be used to approach these current problems, but fusing these models together allows the multimodal model to leverage complimentary information across the different modalities \cite{yhuang}. Huang et al. prove that multimodal learning is better than unimodal based on the intuition that multimodal learning can more accurately estimate the latent space representation \cite{yhuang}. Therefore, multimodal fusion plays a key role in solving the most pressing problems of our time and will only become more important as machine learning (ML) is further integrated into our lives.

The principle behind multimodal systems mimics the way that humans reason about our world. When we draw conclusions from our environment we do not consider solely one type of data, but instead extract any interpretable information from all the data available to us. We apply this same concept to multimodal ML, which builds models that can learn from and inference on multiple types of data. Though standard ML model architectures are typically structured around one type of data input, these unimodal architectures can be fused together in various ways to accept multiple types of data streams concurrently. However, inference latency is often a priority with ML models, especially when deployment to edge devices is necessary. Therefore, the data streams must be carefully chosen based on how much relevant information they provide and how their processing would impact resource consumption. Those selected data streams must also be fused together with consideration for the accuracy and inference latency trade-off. As we show, fusing the data at different points results in varying inference latencies, making some architectures more suitable for edge deployment than others.

We investigate multimodal fusion strategies through the lens of sentiment analysis, a common multimodal fusion problem. We present a hybrid BERT and MobileNetV2 framework that uses both image and text data for early, intermediate, and late fusion strategies. We analyze the trade-off between accuracy and inference latency on these proposed model architectures and show that late fusion yields the best accuracy while early fusion yields lower latency, demonstrating that fusing models earlier is more suitable for edge deployment. We illustrate this trade-off space in Figure~\ref{fig:tradeoff}. A similar trade-off space can be created for any multimodal ML problem, and will be especially useful in domains where inference latency is highly relevant.
\section{Background and Related Works}

\subsection{Multimodal Machine Learning}
Multimodal ML rests on the principles that modalities are heterogeneous, connected, and interact together, providing new insights and yielding higher accuracy when combined \cite{10.1145/3656580}. As the ML systems we build are more intimately integrated with people's lives, such as with the most recent GPT models, there is a demand to interface with models in more intuitive ways, which includes expanding the capability of these models to accept multiple streams of input. However, technical challenges arise in how to best fuse modalities to exploit the interconnectedness between modalities \cite{10.1145/3656580}.

One such challenge is fusing ML models when the architecture of the base models is vastly different. Traditionally, models created for visual processing use a convolutional neural network (CNN) as their base architecture \cite{cnn-review}. In contrast, transformer-based architectures extract textual features better than other model architectures, including CNNs, as is shown by Devlin et al. \cite{bert}. CNN-based and transformer-based architectures have entirely different processing methods, meaning fused models may favor one modality the other as it's features are more accurately extracted.

Another technical challenge is the decision of where in the ML framework the models should be fused together. Models can be fused at different stages in their architecture, often referred to as early, intermediate (or feature-level, or mid-level), and late (or decision-level) fusion \cite{boulahia}. Late fusion allows much of the models to remain the same, fusing at the decision level after feature extraction is complete. This strategy is easier to implement than early strategies, but may fail to fully capture the interconnectedness between the modalities, as processing of the data is almost fully complete in the separate networks before the features are finally fused together. Intermediate fusion happens earlier in the process, within the feature extraction layers. The fused layers are often intertwined with the feature extraction layers of each unimodal model. Finally, early fusion happens even earlier in the architecture. The data may be fused at the data level (before input to the model), after encoding, or after a few layers of feature extraction. The data is fused and the multimodal framework becomes a single fused model. Liang et al. provide an in-depth review of the current state of multi-modal ML in \cite{10.1145/3656580}.

\subsection{Sentiment Analysis}
Sentiment analysis began as a text-based task, powered by a large amount of publicly available opinions on the Internet in the form of posts, tweets, and reviews. It has since become a common multimodal problem \cite{10.1145/3586075}. Visual and audio data, combined with textual data, can provide a more comprehensive picture of the sentiment of the speaker by capturing the non-verbal ques that are so important to speech between humans. The visual data is often an image of the speaker's face, to which facial expression recognition (FER) can be applied. 

Facial expression recognition is a classification problem that seeks to classify a face as expressing a certain emotion. The classes may be specific emotions such as happiness, surprise, sadness, etc. as used in the FER+ dataset \cite{barsoum}. The facial expressions can also be classified into more coarse-grained categories of positive, negative, and neutral \cite{elsayad}. Datasets with fine-grained labels such as FER+ can be reduced to these categories. Depending on the application, a reduced class list may provide a rigorous enough approach, and may in fact be necessary when limited by the classification capabilities of other data streams in the multimodal system. Sajjad et al. provide a current survey of the challenges and applications of FER \cite{sajjad}. In a classical sentiment analysis, the text input data is often written text collected from social media sites or review forums \cite{bashiri}. When considering classification labels for the sentiment of reviews and statements made to social media, a detailed emotional classification such as ``surprise" or ``fear" would be difficult to obtain. Many text-only sentiment analysis approaches focus on positive and negative classification. This decision is also informed by the purpose of the development of textual sentiment analysis methods, which are often used to obtain consumer or political data where a general sentiment is sufficient for the interested party \cite{bashiri}.

\subsubsection{Multimodal Sentiment Analysis}
In multimodal sentiment analysis, the FER model and textual sentiment analysis model are combined into one uniform model. Considering classification labels that work for both image and text data, classification labels of positive, negative, and neutral to varying degrees are often favored over the emotional labels that can be used in FER. Bashiri and Naderi release 22 datasets with sentiments labeled as positive or negative \cite{bashiri}.

Multimodal sentiment analysis also frequently includes audio recordings of the speaker. Audio data can capture important sentiment indicators, such as tone and volume, that are not revealed in visual or textual data \cite{10.1145/3586075}. In some domains this audio data may be highly relevant, but in sentiment analysis it is shown by \cite{10.1145/3586075} to have a marginal impact on accuracy compared to image and textual data. For this reason, we will not consider audio data as an input stream in this work.

\subsubsection{Fusion Strategies}
Much of the recent research in multimodal sentiment analysis focuses on the development of novel fusion strategies. Majumder et al. focus on the contextual relationships between the different modalities by first fusing each modality (textual, visual, audio) with each other modality, and then fusing the resulting dual-modal layers with each other before predicting the final output \cite{majumder}. Their accuracy on the CMU-MOSI dataset for the combined textual and visual modalities was 79.3\% \cite{majumder}. Huang et al. focus solely on fusing textual and visual data with their Deep Multimodal Attentive Fusion model \cite{HUANG201926}. The authors describe three base model architectures: a visual attention model, a semantic attention model, and multimodal attention model which is structured the same as the visual and semantic models, but fuses the networks at an intermediate stage. They provide results for the visual and semantic models, their multimodal attention model, and a late fusion version of all three models \cite{HUANG201926}. On the Flickr dataset, which they relabeled, they show that the fused models reach higher accuracy than the unimodal models \cite{HUANG201926}. Huddar et al. also focus on attention with their multimodal contextual fusion strategy \cite{huddar}. And, like Majumder et al., focus on the fusion of two modalities before fusing the resulting bimodal features together \cite{huddar}. They use a bidirectional LSTM with an attention model to capture contextual information and similarities between information of different modalities \cite{huddar}. Their proposed models score between 78.5-79.92\% accuracy for the combined textual and visual models on the CMU-MOSI dataset \cite{huddar}.
\section{Methodologies}
We discuss in this section the methodologies used to create the proposed model architectures, including the dataset used for training and the architectural modifications made to the base BERT, MobileNetV2 and ViT architectures.

\subsection{Dataset}
The CMU-MOSI dataset is publicly available and widely applied to multimodal research applications \cite{mosi}. It contains synchronized video clips, spoken phrases, and audio clips of movie reviews. Each of the 2199 video clips shows a single speaker giving their opinion to the camera. The annotations designate the speaker as feeling very negative (-3) to feeling very positive (3) with 0 being neutral. We reduce the labels to negative and non-negative (positive plus neutral) to compute binary accuracy. We also divided the video clips into frames and used the middle frame from the sequence instead of the full clip.

\subsection{Data Streams}
We fuse together the textual and visual data provided by the CMU-MOSI dataset to build a complete picture of the sentiment of the speaker. As previously discussed, the audio data does not contribute significantly to the accuracy of multimodal sentiment analysis systems, and is therefore left off. We also want to note that trade-off between accuracy and inference latency with each added data modality. Including auditory data as a third data stream in our multimodal system may marginally increase the accuracy, but it would negatively impact the inference speed of the model, as the architecture would need to be expanded to process this separate data stream. The added data stream would have a greater impact on later fusion methods than earlier, as more processing would be devoted to auditory data alone in later fusion methods. This impact on inference latency is another reason that auditory data was not considered, even if it may give marginal accuracy benefits.

\subsection{Architectural Modifications}
We present three model architectures at varying levels of fusion, all based on a combination of BERT and MobileNetV2 \cite{bert, mobilenetv2}. The model architectures are fused at three different stages of the overall architecture (late, intermediate, early), providing different levels of accuracy and latency. We show that the level of fusion is important to consider when using resource-constrained devices for inference, as fusion at lower levels will provide a higher accuracy while fusion at higher levels provides lower latency. Furthermore, we provide three additional model architectures which use the same fusion strategies, but replace MobileNetV2 with ViT. These models show the same accuracy-latency trade-off as the BERT-MobileNetV2 models.

\subsubsection{Late Fusion}
First, we fine-tune the pretrained bert-base-uncased BERT model and MobileNetV2 (pretrained google/mobilenet\_v2\_1.4\_224) independently on a subset of the CMU-MOSI dataset \cite{bertmodel, mobilenetv2model}. We then use these trained base models for our late fusion strategy.

\begin{figure}
    \centering
    \includegraphics[width=\linewidth]{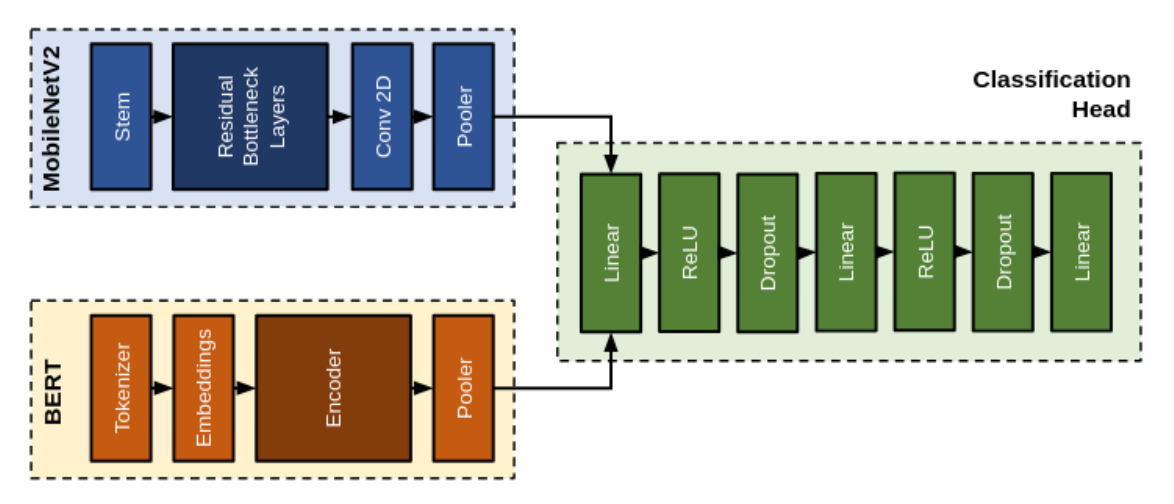}
    \caption{Architecture of BERT and MobileNetV2 fused at the late stage. The simplified structures of BERT and MobileNetV2 are shown. Their individual classification heads are removed and instead the outputs from the pooling functions of models are passed directly to a unified classification head.}
    \label{fig:late}
\end{figure}

The late fusion model fuses the fine-tuned BERT and MobileNetV2 models together. Both image and textual data are input to the model, which sends the textual data through the frozen BERT model and image data through the frozen MobileNetV2 model. Instead of their independent classification heads, a new classification head is created that makes a decision considering the output of both models. This combined model is fine-tuned with another subset of the CMU-MOSI dataset. We then show that the accuracy of the fused model is higher than either model independently. This shows that the combined modalities provide a more holistic view of the speaker’s emotion which can be captured through a multimodal approach. The late fusion architecture is shown in Figure~\ref{fig:late}.

\begin{figure*}
    \centering
    \includegraphics[width=\linewidth]{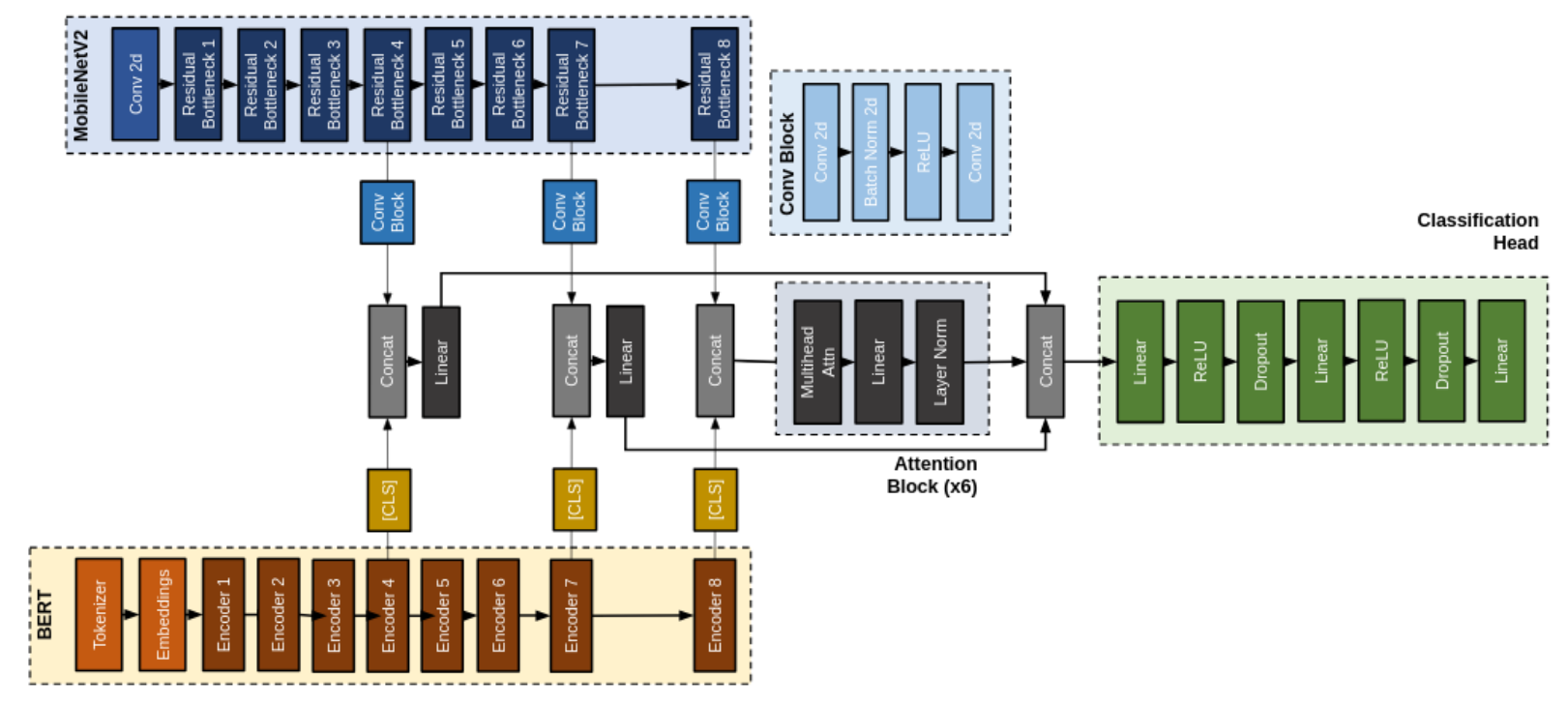}
    \caption{Architecture diagram of BERT and MobileNetV2 fused at the intermediate level. Features are extracted from both unimodal models at separate points and concatenated before being passed to a final classification head.}
    \label{fig:mid}
\end{figure*}

\begin{figure*}
    \centering
    \includegraphics[width=\linewidth]{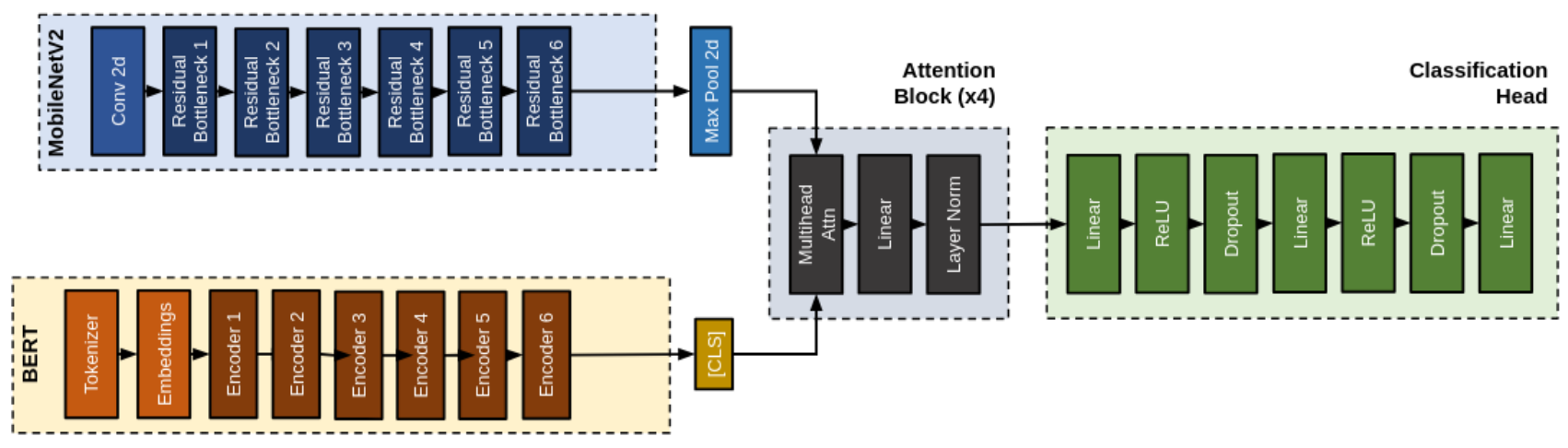}
    \caption{Architecture diagram of BERT and MobileNetV2 fused at the early stage. The base models are cut off after six feature extraction layers and their output is combined and passed to a unified model which uses attention blocks to extract more features before passing to a classification head.}
    \label{fig:early}
\end{figure*}

\subsubsection{Intermediate Fusion}
The intermediate fusion is done in the encoder layers of the BERT model and the intermediate layers of the MobileNetV2 model. The original BERT encoder has 12 layers and the MobileNetV2 model has 16 residual bottleneck layers \cite{bert, mobilenetv2}. The proposed model architecture uses the first eight encoder and intermediate layers of BERT and MobileNetV2, respectively. The rest of the original model architectures is not used. In these eight encoder and residual bottleneck layers, the outputs of three selected layers at the same depth are concatenated together. These joined features from the first two fused layers are passed through a linear function and the joined features from from the last selected layer are passed through an attention block before all three of the joined features are concatenated together. This output capturing features extracted at different depths in the models is passed through a classification head. Figure~\ref{fig:mid} shows this proposed architecture.

Because textual features are highly relevant for our application, the decision of which BERT layers to use for feature extraction has significant impact on the overall accuracy of the model. Since BERT also has a greater impact on the inference latency, the BERT layers we use must be carefully chosen. It has been shown that BERT's encoder layers process linguistic information hierarchically, with early (1-4) and middle (5-8) layers capturing simple and complex syntactic features, respectively, and late layers (9-12) capturing semantic features and longer-range dependencies \cite{bert-probe}. We hypothesize that, while helpful, the sentiment of a statement needs less detailed semantic understanding over a long range than other language processing tasks since a positive or negative sentiment can often be inferred by the presence of emotionally-charged words. We therefore consider only the first eight layers for our intermediate fusion model. Since the type of processing done by BERT's encoding layers is quantitatively observed to generally follow groupings of four, layer four may represent completion of a first stage of processing and layer eight may represent the end of stage two \cite{bert-probe-quant}. The features extracted at the end of these stages, captured by output of encoder layers four and eight, are then forwarded to the fused layers. The output of encoder layer seven is also used in the fused model as the deeper layers will be the most feature-rich.

The hierarchical nature of feature extraction for MobileNetV2 has not been as rigorously examined, but it can be inferred from the network architecture that deeper layers extract more complex visual features \cite{mobilenetv2}. Lacking this interpretability, features extracted from MobileNetV2's residual bottleneck layers at the same level of BERT's encoder layers were used.

\subsubsection{Early Fusion}
The proposed early fusion method combines the two unimodal models into one model early into the architecture while preserving valuable feature extraction layers of both base models. The data is processed separately in the two base models up to the sixth feature extraction layer. After this layer, the results are joined and passed through four attention blocks to extract joined features. Our ablation study below tests the accuracies of different numbers of attention blocks. The same classification head used in late and intermediate fusion is used. Figure~\ref{fig:early} shows the proposed architecture.

When combining models of vastly different architectures, like the CNN MobileNetV2 with the transformer-based BERT, it can be difficult to choose what kind of architecture to merge the models into. We have already seen that the textual features are more relevant to the sentiment classification than the image features for this dataset. We created the early fusion model to leverage this strength by using an attention-based architecture to extract the textual features.

\subsection{ViT Fusion}
Vision transformers (ViTs) have been shown to match or pass CNNs in terms of accuracy for processing image data \cite{vit}. We also fuse the BERT model with a ViT model in the place of the MobileNetV2 model to evaluate how the vision model affects accuracy and latency results. For late fusion, the base ViT model is pretrained (google/vit-base-patch16-224)\cite{vitmodel}. ViT has the same number of encoder layers as BERT, so features were extracted at the same layers as BERT \cite{vit}.

\subsection{Training}
The base MobileNetV2 model was fine-tuned for 150 epochs. We used stochastic gradient descent for optimization with a learning rate of 1e-4, momentum of 0.9 and weight decay of 1e-4. The BERT base model was fine-tuned for 12 epochs. We used the Adam optimizer with a learning rate of 1e-5, beta coefficients 0.9 and 0.999 and epsilon of 1e-8. For the late fusion model we trained for 100 epochs. The weights of the BERT model and MobileNetV2 model were frozen during training of the late fusion model. The learning rate used was 1e-3 with a momentum of 0.9 and a weight decay of 1e-4. Intermediate and early fusion do not use fine-tuned base models due to the extensive changes that are made within the architecture of the models. The intermediate fusion and early fusion models use a stochastic gradient descent optimizer with learning rate of 1e-4, momentum of 0.9, and weight decay 1e-4. The multihead attention layer in the attention block used in the intermediate fusion and early fusion model has 4 heads. A batch size of 32 was used for all models. Training for the ViT-fused models uses the same hyperparameters and optimizer as the MobileNetV2-fused models.
\section{Performance Trade-Offs}

\subsection{Accuracy}
As shown in Table~\ref{tab:accuracy_all}, the accuracy of our best model, the late fusion model, is comparable to the state-of-the-arts models on this dataset.

\begin{table}[t]
\centering
\caption{Accuracy performance on the CMU-MOSI dataset for the most recent models according to the leaderboard in the CMU-MultimodalSDK repository \cite{leaderboard}. BA denote binary accuracy. F1 denotes f1-score. The human performance is reported in \cite{10.1145/3656580}.}
\label{tab:accuracy_all}
\begin{tabular}{lcc}
\toprule
\textbf{Model} & \textbf{BA} &\textbf{F1} \\
\midrule
ARGF & - & 81.4 \\ 
MulT (ACL20) & - & 83 \\
TFN (B) (EMNLP17) & - & 80.8 \\ 	
LMF (B) (ACL18) & - & 82.5 \\ 	
MFM (B) (ICLR19) & - & 81.7 \\
ICCN (B) & - & 83 \\
FMT & 81.5 & 83.5 \\
MISA (B) &	81.8 & 83.4 \\	
MAG-BERT (ACL20) &	84.2 & 84.1 \\
MAG-XLNet (ACL20) & 85.7 & 85.6 \\
Human & 85.7 & 87.5 \\
\textit{Late Fusion (Ours)} & 84.3 & - \\
\bottomrule
\end{tabular}
\end{table}

Table~\ref{table:accuracy-results} shows the accuracy of each proposed model on a subset of the CMU-MOSI dataset used for testing. It can be seen that while the accuracy of the base MobileNetV2 model on images only is low, it can be leveraged in combination with BERT to receive a fused accuracy that surpasses both base models. The late fusion model has the highest accuracy of the fused models, with early fusion being the least accurate and intermediate fusion in the middle. This is likely due to the amount of specialized feature extraction layers that are reduced as fusion occurs earlier in the architecture. Both the intermediate and early fusion models are not as accurate as the BERT model is independently, but, as we will discuss, the inference latency gives these models some benefit in the trade-off space.

\begin{table}[t]
\centering
\caption{Accuracy results on a CMU-MOSI test subset. The modality column includes V for visual and T for textual. BA denotes binary accuracy. M denotes MobileNetV2-fused models and V denotes ViT-fused models.}
\label{table:accuracy-results}
\begin{tabular}{lccc}
\toprule
\textbf{Model} & \textbf{Modality} & \textbf{BA (\%) (M)} & \textbf{BA (\%) (V)} \\
\midrule
MobileNetV2         & V            & 43.22 & - \\
ViT                 & V            & - & 50.17 \\
BERT                & T           & 80.20 & - \\
Late Fusion         & T + V  & 84.25 & 83.62 \\
Inter. Fusion & T + V  & 72.40 & 69.90 \\
Early Fusion        & T + V  & 67.89 & 66.59 \\
\bottomrule
\end{tabular}
\end{table}

We found the accuracy of the ViT-fused model to be comparable to the accuracy of the MobileNetV2-fused model, with the MobileNetv2-fused model consistently performing slightly better. This points to the language model having a more significant impact on the results. Because the base ViT model performed better than the base-MobileNetV2 model, the higher accuracy of MobileNetV2-based models indicates that the MobileNetV2 model fuses with the BERT model in a more meaningful way for this task.

\subsection{Ablation Study}
The early fusion model includes a custom unimodal attention network with four attention blocks. The number of attention blocks was chosen by training four different BERT-MobileNetV2 early fusion models with different numbers of attention blocks. A four-block model and eight-block model yielded the best accuracy. Since the accuracy is comparable and we need to keep latency in mind, we selected the smaller 4-block network over the 8-block network. The BERT-ViT model follows the same structure. Table~\ref{table:early-ablation} shows the accuracies of the different numbers of attention blocks.
\begin{table}[t]
\centering
\caption{Ablation study on the number of attention blocks added to the early fusion model.}
\label{table:early-ablation}
\begin{tabular}{llc}
\toprule
\textbf{\# of Attn. Blocks} & \textbf{Accuracy (\%)} \\
\midrule
2 & 62.37 \\
4 & 67.89 \\
6 & 61.38  \\
8 & 69.09 \\
\bottomrule
\end{tabular}
\end{table}

\subsubsection{Inference Latency}
The proposed models were deployed to a Jetson Orin AGX. These models were converted from PyTorch to ONNX format. We show that, while the late fusion model did provide the most accuracy, it comes at the cost of a higher inference latency. The intermediate and early fusion models have lower inference latencies, with early fusion providing the lowest. We show that while these models are less accurate than the late fusion model or even then the unimodal BERT, they come at a lower inference latency cost making them more applicable to some resource-constrained applications.

Table~\ref{table:latency-results} shows the inference latency with ONNX runtime, excluding any optimizations added by TensorRT. When comparing our proposed models with the base models we use ONNX runtime because we expect that TensorRT will not apply the same level of optimizations to our custom models as it can to more traditional models such as BERT, MobileNetV2, and ViT.

To our knowledge this is the first reporting of inference latency metrics for the CMU-MOSI dataset. The models reporting leading accuracy in the CMU-MOSI score board are MAG-XLNET and MAG-BERT as previously discussed \cite{rahman}. While their inference latency is not reported, our latency is likely comparable to MAG-BERT since they also use BERT for textual processing.

\begin{table}[t]
\centering
\caption{Inference latency (in seconds) for the proposed models on a NVIDIA Jetson Orin AGX. ONNX models are used for these results. We also provide accuracy results here to illustrate the accuracy-latency trade-off.}
\label{table:latency-results}
\begin{tabular}{lcc}
\toprule
\textbf{Model} & \textbf{Accuracy (\%)} & \textbf{Latency (ms) (M)} \\
\midrule
Late Fusion   & 84.25 & 21.6\\
BERT          &   80.20   & 18.6  \\
Inter. Fusion &72.40 & 13.5\\
Early Fusion  &   67.89  & 11.4 \\
MobileNetV2   &  43.22    &  	3.0 \\
\bottomrule
\end{tabular}
\end{table}
The ViT-fused latency results are shown in Table~\ref{table:latency-results-trt}. Understandably, the ViT-fused models had a slower inference time than the MobileNetV2-fused models, since MobileNetV2's independent inference latency of 0.98 milliseconds (with TensorRT optimization) is less than that of ViT at 6.23 milliseconds (with TensorRT optimization).
\begin{table}[t]
\centering
\caption{Inference latency (in seconds) for the ViT-fused models on a NVIDIA Jetson Orin AGX. The models are optimized with TensorRT. M denotes MobileNetV2-fused models and V denotes ViT-fused models.}
\label{table:latency-results-trt}
\begin{tabular}{lcc}
\toprule
\textbf{Model} & \textbf{Latency (ms) (M)} & \textbf{Latency (ms) (V)} \\
\midrule
Late Fusion     & 13.6909 & 18.7897 \\
Inter. Fusion & 9.42474 & 13.1855 \\
Early Fusion       & 7.32538 & 10.951 \\
\bottomrule
\end{tabular}
\end{table}
As we have shown, earlier fusion strategies can provide lower inference latency, which is important when creating multimodal systems for edge deployment, where decisions must be made with reduced resources in a timely manner.
\section{Conclusion}
We have shown that considering the accuracy and latency trade-off when creating multimodal systems is critical for systems where speed of service is important. We provide three proposed models that show how the trade-off space is impacted by fusion at different stages in the model architecture. Figure~\ref{fig:tradeoff} plots this trade-off space. We show that, for our case, late fusion is often more accurate due to feature extraction that is specialized for the input data. However, we also show that earlier fusion will provide a faster inference speed. We argue that this trade-off should be considered in any multimodal machine learning strategy. As multimodal systems evolve, it is critical that we keep inference latency in mind.

\subsection{Future Work}
While our work provides a basis for exploring accuracy and inference latency trade-off in multimodal systems, this work may be expanded in many ways.

One such expansion is to fuse different model architectures together. We used two different vision models, but we may use a different language model or yet another type of vision model. We may also explore differently-sized BERT models to observe the difference in performance across different edge devices.

This work may also be expanded by including audio data and quantifying the performance trade-off of adding this extra modality. The impact to inference latency would depend heavily on how early the auditory data stream could be merged with another data stream. This relates to the compatibility of model architectures, as we discuss in the Early Fusion subsection.

We also note that the existence of a performance trade-off between these three proposed models implies other, differently fused models that fill empty spaces in the accuracy-latency trade-off space. It is possible to build these models by hand, but it may also be useful to represent this problem as a multi-objective optimization problem.

Software such as Optuna already allows for an automated neural architecture search that optimizes for accuracy by tuning hyperparameters like learning rate and weight decay \cite{optuna_2019}. A second optimization objective of inference latency automates the population of the performance trade-off space discussed in this work, potentially yielding further insight into how model architecture affects accuracy and latency, while also providing a framework for dynamically generating these different multimodal architectures. The challenge is in encoding a vector-representation of the model architecture to enable dynamic generation of varying network connections including different concatenation strategies and feature extraction layers.

Many multimodal works may benefit from providing multiple models in different positions in their performance trade-off space, especially when inference speed is an important consideration.

{
    \small
    \bibliographystyle{ieeenat_fullname}
    \bibliography{ref}

@String(NIPS= {Adv. Neural Inform. Process. Syst.})

@String(NIPS  = {NeurIPS})

@article{mosi,
  	title="Mosi: multimodal corpus of sentiment intensity and subjectivity analysis in online opinion videos.",
	author="Zadeh, Amir and Zellers, Rown and Pincus, Eli and Morency, Louis-Philippe",
	publisher="arXiv preprint arXiv:1606.06259",
	year="2016"
}

@article{10.1145/3586075,
author = {Das, Ringki and Singh, Thoudam Doren},
title = {Multimodal Sentiment Analysis: A Survey of Methods, Trends, and Challenges},
year = {2023},
issue_date = {December 2023},
publisher = {Association for Computing Machinery},
address = {New York, NY, USA},
volume = {55},
number = {13s},
issn = {0360-0300},
url = {https://doi.org/10.1145/3586075},
doi = {10.1145/3586075},
journal = {ACM Comput. Surv.},
month = jul,
articleno = {270},
numpages = {38}
}

@article{huddar,
	title="Attention-based multimodal contextual fusion for sentiment and emotion classification using bidirectional LSTM",
	author="Huddar, M.G. and Sannakki, S.S. and Rajpurohit, V.S.",
	year="2021",
	issue_date="11 January 2021",
	publisher="Multimedia Tools and Applications",
	volume="80",
	url="https://doi.org/10.1007/s11042-020-10285-x",
}

@online{leaderboard,
	author="CMU-MultiComp-Lab",
	title="CMU-Multimodal Opinion Sentiment Intensity (MOSI) dataset Scoreboard",
	year="2023",
	url="https://github.com/CMU-MultiComp-Lab/CMU-MultimodalSDK/tree/main/mmsdk/mmdatasdk/dataset/standard_datasets/CMU_MOSI",
	urldate="2025-09-14"
}

@article{majumder,
	title="Multimodal Sentiment Analysis using Hierarchical Fusion with Context Modeling",
	author="Majumder, N. and Hazarika, D. and Gelbukh, A. and Cambria, E. and Poria, S.",
	year="2018",
	publisher="Knowledge-Based Systems",
	volume="161",
	issue_date="1 December 2018",
	url="https://doi.org/10.1016/j.knosys.2018.07.041"
}

@article{HUANG201926,
title = {Image–text sentiment analysis via deep multimodal attentive fusion},
journal = {Knowledge-Based Systems},
volume = {167},
pages = {26-37},
year = {2019},
issn = {0950-7051},
doi = {https://doi.org/10.1016/j.knosys.2019.01.019},
url = {https://www.sciencedirect.com/science/article/pii/S095070511930019X},
author = {Feiran Huang and Xiaoming Zhang and Zhonghua Zhao and Jie Xu and Zhoujun Li}
}

@article{rahman,
	title="Integrating Multimodal Information in Large Pretrained Transformers",
	author="Rahman, Wasifur and Hasan, Md Kamrul and Lee, Sangwu and Zadeh, Amir and Mao, Chengfeng and Morency, Louis-Philippe and Hoque, Ehsan",
	publisher="Proc Conf Assoc Comput Linguist Meet",
	year="2020",
	issue_date="July 2020",
	doi="10.18653/v1/2020.acl-main.214"
}

@article{10.1145/3656580,
author = {Liang, Paul Pu and Zadeh, Amir and Morency, Louis-Philippe},
title = {Foundations \& Trends in Multimodal Machine Learning: Principles, Challenges, and Open Questions},
year = {2024},
issue_date = {October 2024},
publisher = {Association for Computing Machinery},
address = {New York, NY, USA},
volume = {56},
number = {10},
issn = {0360-0300},
url = {https://doi.org/10.1145/3656580},
doi = {10.1145/3656580},
journal = {ACM Comput. Surv.},
month = jun,
articleno = {264},
numpages = {42}
}

@article{boulahia,
	title="Early, intermediate and late fusion strategies for robust deep learning-based multimodal action recognition",
	author="Boulahia, Said Yacine and Amamra, Abdenour and Madi, Mohamed Ridha and Daikh, Said",
	year="2021",
	publisher="Machine Vision and Applications",
	volume="32",
	issue_date="30 September 2021"
}

@online{bertmodel,
	author="Hugging Face",
	title="BERT base model (uncased)",
	url="https://huggingface.co/google-bert/bert-base-uncased",
	urldate="2025-09-15"
}

@online{mobilenetv2model,
	author="Hugging Face",
	title="MobileNetV2",
	url="https://huggingface.co/google/mobilenet_v2_1.4_224",
	urldate="2025-09-15"
}

@article{sajjad,
title = {A comprehensive survey on deep facial expression recognition: challenges, applications, and future guidelines},
journal = {Alexandria Engineering Journal},
volume = {68},
pages = {817-840},
year = {2023},
issn = {1110-0168},
doi = {https://doi.org/10.1016/j.aej.2023.01.017},
url = {https://www.sciencedirect.com/science/article/pii/S1110016823000327},
author = {Muhammad Sajjad and Fath U Min Ullah and Mohib Ullah and Georgia Christodoulou and Faouzi {Alaya Cheikh} and Mohammad Hijji and Khan Muhammad and Joel J.P.C. Rodrigues}
}

@article{khuang,
  title={Multi-modal Sensor Fusion for Auto Driving Perception: A Survey},
  author={Keli Huang and Botian Shi and Xiang Li and Xin Li and Siyuan Huang and Yikang Li},
  journal={ArXiv},
  year={2022},
  volume={abs/2202.02703},
  url={https://api.semanticscholar.org/CorpusID:246634264}
}

@ARTICLE{ghosh,
       author = {{Ghosh}, Akash and {Acharya}, Arkadeep and {Saha}, Sriparna and {Jain}, Vinija and {Chadha}, Aman},
        title = "{Exploring the Frontier of Vision-Language Models: A Survey of Current Methodologies and Future Directions}",
      journal = {arXiv e-prints},
         year = 2024,
        month = feb,
          eid = {arXiv:2404.07214},
        pages = {arXiv:2404.07214},
          doi = {10.48550/arXiv.2404.07214},
archivePrefix = {arXiv},
       eprint = {2404.07214},
 primaryClass = {cs.CV},
       adsurl = {https://ui.adsabs.harvard.edu/abs/2024arXiv240407214G}
}

@inproceedings{yhuang,
author = {Huang, Yu and Du, Chenzhuang and Xue, Zihui and Chen, Xuanyao and Zhao, Hang and Huang, Longbo},
title = {What makes multi-modal learning better than single (provably)},
year = {2021},
isbn = {9781713845393},
publisher = {Curran Associates Inc.},
address = {Red Hook, NY, USA},
booktitle = {Proceedings of the 35th International Conference on Neural Information Processing Systems},
articleno = {837},
numpages = {13},
series = {NIPS '21}
}

@article{cnn-review,
    authors="Zhao, Xia and Zhang, Yufei and Han, Xuming and Deveci, Muhammet and Parmar, Milan",
    title="A review of convolutional neural networks in computer vision",
    year="2024",
    issue_date="23 March 2024",
    journal="Artif Intell Rev",
    volume="57",
    articleno="99",
    url="https://link.springer.com/article/10.1007/s10462-024-10721-6",
    doi="https://doi.org/10.1007/s10462-024-10721-6"
}

@article{bert,
    title="BERT: Pre-training of Deep Bidirectional Transformers for
Language Understanding",
    authors="Devlin, Jacob and Chang, Ming-Wei and Lee, Kenton and Toutanova, Kristina",
    journal="Association for Computational Linguistics",
    year="2019",
    issue_date="7 June 2019",
    url="https://arxiv.org/abs/1810.04805"
}

@inproceedings{barsoum,
author = {Barsoum, Emad and Zhang, Cha and Ferrer, Cristian Canton and Zhang, Zhengyou},
title = {Training deep networks for facial expression recognition with crowd-sourced label distribution},
year = {2016},
isbn = {9781450345569},
publisher = {Association for Computing Machinery},
address = {New York, NY, USA},
url = {https://doi.org/10.1145/2993148.2993165},
doi = {10.1145/2993148.2993165},
booktitle = {Proceedings of the 18th ACM International Conference on Multimodal Interaction},
pages = {279–283},
numpages = {5},
location = {Tokyo, Japan},
series = {ICMI '16}
}

@article{elsayad,
    title="An automatic improved facial expression recognition for masked faces",
    authors="ELsayed, Yasmeen and ELSayed, Ashraf and Abdou, Mohamed A.",
    journal="Neural Computing and Applications",
    volume="35",
    year="2023",
    issue_date="July 2023",
    url="https://link.springer.com/article/10.1007/s00521-023-08498-w#citeas",
    doi="https://doi.org/10.1007/s00521-023-08498-w"
}

@article{bashiri,
    authors="H. Bashiri and H. Naderi",
    title="Comprehensive review and comparative analysis of transformer models in sentiment analysis",
    year="2024",
    publisher="Knowl Inf Syst",
    volume="66",
    issue_date="30 September 2021"
}

@article{vit,
  title={An Image is Worth 16x16 Words: Transformers for Image Recognition at Scale},
  author={Alexey Dosovitskiy and Lucas Beyer and Alexander Kolesnikov and Dirk Weissenborn and Xiaohua Zhai and Thomas Unterthiner and Mostafa Dehghani and Matthias Minderer and Georg Heigold and Sylvain Gelly and Jakob Uszkoreit and Neil Houlsby},
  journal={ArXiv},
  year={2020},
  volume={abs/2010.11929},
  url={https://api.semanticscholar.org/CorpusID:225039882}
}

@article{mobilenetv2,
  title={MobileNetV2: Inverted Residuals and Linear Bottlenecks},
  author={Mark Sandler and Andrew G. Howard and Menglong Zhu and Andrey Zhmoginov and Liang-Chieh Chen},
  journal={2018 IEEE/CVF Conference on Computer Vision and Pattern Recognition},
  year={2018},
  pages={4510-4520},
  url={https://api.semanticscholar.org/CorpusID:4555207}
}

@inproceedings{optuna_2019,
    title={Optuna: A Next-generation Hyperparameter Optimization Framework},
    author={Akiba, Takuya and Sano, Shotaro and Yanase, Toshihiko and Ohta, Takeru and Koyama, Masanori},
    booktitle={Proceedings of the 25th {ACM} {SIGKDD} International Conference on Knowledge Discovery and Data Mining},
    year={2019}
}

@inproceedings{bert-probe,
  title={What does BERT learn about the structure of language?},
  author={Jawahar, Ganesh and Sagot, Beno{\^\i}t and Seddah, Djam{\'e}},
  booktitle={ACL 2019-57th Annual Meeting of the Association for Computational Linguistics},
  year={2019}
}

@article{bert-probe-quant,
  title={BERT rediscovers the classical NLP pipeline},
  author={Tenney, Ian and Das, Dipanjan and Pavlick, Ellie},
  journal={arXiv preprint arXiv:1905.05950},
  year={2019}
}

@online{vitmodel,
	author="Hugging Face",
	title="Vision Transformer (base-sized model)",
	url="https://huggingface.co/google/vit-base-patch16-224",
	urldate="2025-11-12"
}
}

\end{document}